\documentclass[journal]{IEEEtran}
\usepackage{bm}
\usepackage{times}
\usepackage{amsmath}
\usepackage{graphicx}
\usepackage{warpcol}
\usepackage[ruled,vlined]{algorithm2e}
\usepackage{amssymb}
\usepackage{booktabs}
\usepackage{multirow}
\usepackage[pagebackref=false,breaklinks=true,letterpaper=true,colorlinks,bookmarks=false,citecolor={blue},linkcolor={blue}]{hyperref}
\usepackage{soul}

\usepackage{cite}

\ifCLASSINFOpdf
\else
\fi
\begin{document}
%
\title{Pseudo-LiDAR Based Road Detection}

%
%
%

\author{Libo~Sun, 
        Haokui~Zhang, 
        and Wei Yin 
\thanks{L. Sun and W. Yin are with The University of Adelaide, Australia (e-mail: libo.sun@adelaide.edu.au; wei.yin@adelaide.edu.au). 

H. Zhang is with School of Computer Science, Northwestern Polytechnical University, Xi'an, and also with Intellifusion, Shenzhen, China (e-mail: hkzhang1991@mail.nwpu.edu.cn).

Part of this work was carried out when H. Zhang was visiting The University of Adelaide.
L. Sun and H. Zhang contributed equally to this work.}
}

\maketitle

\begin{abstract}
Road detection is a critically important task for self-driving cars. By employing LiDAR data, recent works have significantly improved the accuracy of road detection. However, relying on LiDAR sensors limits the application of those methods when only cameras are available. In this paper, we propose a novel road detection approach with RGB images being the only input. Specifically, we exploit pseudo-LiDAR using depth estimation and propose a feature fusion network in which RGB images and learned depth information are fused for improved road detection. To optimize the network architecture and improve the efficiency of our network, we propose a method to search for the information propagation paths. Finally, to reduce the computational cost, we design a modality distillation strategy to avoid using depth estimation networks during inference. The resulting model eliminates the reliance on LiDAR sensors and
achieves state-of-the-art performance on two challenging benchmarks, KITTI and R2D.
\end{abstract}


%
\IEEEpeerreviewmaketitle

\section{Introduction}
%
%
%
%
\IEEEPARstart{R}{oad} detection is indispensable in the field of autonomous driving and advanced driver assistance systems (ADAS). As a critical component for autonomous driving, road detection aims to detect the drivable areas for autonomous vehicles to make reliable decisions. Due to the great importance of road detection, a large number of related works~\cite{chen2017rbnet, han2018semisupervised, he2004color, alvarez2014combining, chen2019progressive, fan2020sne} have been proposed in recent years. More recently, with the help of deep convolutional neural networks (DCNNs)~\cite{krizhevsky2012imagenet, lecun2015deep}, the performance of road detection has been significantly improved.

Introducing DCNNs can obtain accurate road detection results, but it fails when facing some extreme situations, such as ambiguous boundaries and shadow areas. To solve these problems and to improve road detection accuracy, a large number of LiDAR-based
works~\cite{caltagirone2019lidar2, chen2019progressive, fan2020sne} have been proposed. These works have demonstrated that introducing LiDAR information can significantly increase the road detection accuracy. However, it leads to another problem that LiDAR data needs to be obtained by using expensive devices,
which increases the cost on one hand and limits 
their application when only cameras are available on the other hand.

\begin{figure}[t]
\begin{center}
\includegraphics[width=3.4in]{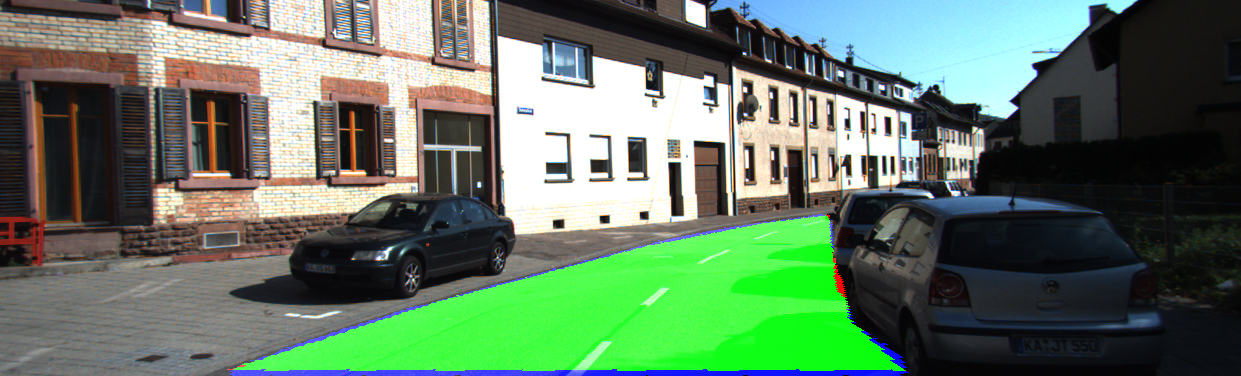}
\\ \hspace*{\fill} \\
\includegraphics[width=3.4in]{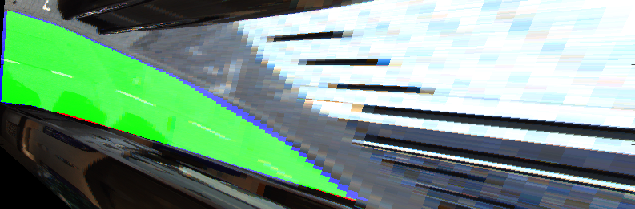}
\end{center}
  \caption{Road detection results by using our method on the KITTI dataset~\cite{fritsch2013new}. The top and bottom parts of the figure show the results from a perspective view and an aerial view respectively. Red represents false negatives, blue areas correspond to false positives, and green denotes true positives.}
 \label{fig:first_page}
\end{figure}

\begin{figure*}[t]
\begin{center}
\includegraphics[width=7.0 in]{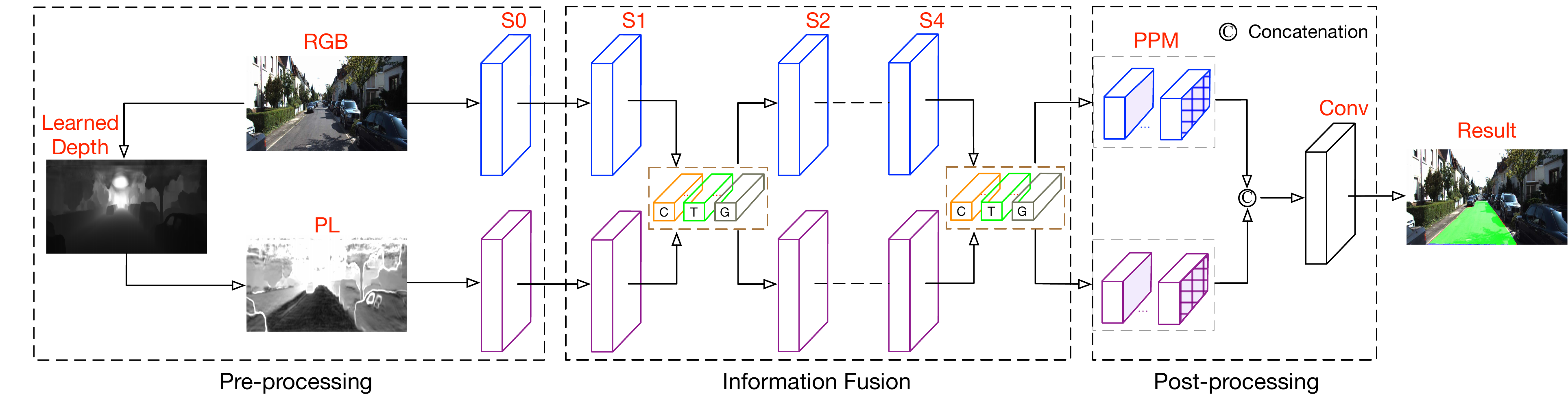}
\end{center}
\caption{The overall pipeline of the proposed method. The whole framework consists of three sub-components, i.e., pre-processing, information fusion, and post-processing.
The residual blocks in ResNet~\cite{he2016deep} are denoted as S0 to S4, and the RGB and depth features are fused in a CTG module (see Fig.~\ref{fig:fusion_component}) from S1 to S4. The RGB and pseudo-LiDAR (PL) information are combined to obtain the final prediction result after the pyramid pooling module (PPM)~\cite{zhao2017pyramid}. Different from the information fusion approach used in LiDAR-based networks~\cite{chen2019progressive, fan2020sne}, two branches exchange their information in the CTG
modules and the searched information propagation paths (see Fig.~\ref{fig: IPPS}) are added to compose the final network.
}
\label{fig:network_overview}
\end{figure*}

In order to address the above limitations, we
propose a new road detection method that does not rely on LiDAR data and can achieve performance comparable to current LiDAR-based methods.
Firstly, 
motivated by the success of using pseudo-LiDAR 
in other autonomous driving related tasks~\cite{wang2019pseudo, weng2019monocular, you2019pseudo}, we propose a pseudo-LiDAR  based road detection framework, in which we use pseudo-LiDAR (PL) learned from RGB images to replace LiDAR.

We further introduce a pseudo-LiDAR based information fusion network (PLIFNet) to synthesize both RGB and pseudo-LiDAR data and mutually update the two input sources. This is different from previous LiDAR-based networks~\cite{chen2019progressive, fan2020sne} that only use the LiDAR branch to update the RGB branch in a single direction.
Secondly,
the current road detection networks~\cite{chen2019progressive,fan2020sne} are manually designed, which are not conducive to the use of all information. To address this issue, we propose an information propagation path search (IPPS) method to
automatically tune and optimize the 
PLIFNet architecture. The optimized PLIFNet can obtain state-of-the-art performance on both the KITTI and R2D datasets without using LiDAR data (see Fig.~\ref{fig:first_page} for an example). 
Lastly, to reduce the computational cost
of inference
, we propose a modality distillation method for the road detection task. Particularly, a new modality distillation loss is introduced to learn expressive features of different granularities from pixel inputs to patches and holistic images.
By doing so, it enables our framework to obtain accurate prediction results without using computationally costly depth networks during inference.

Our main contributions can be summarized as follows:
\begin{itemize}
\item We propose a pseudo-LiDAR based information fusion framework, in which we use pseudo-LiDAR learned from RGB images to replace LiDAR and introduce a new information fusion network, called PLIFNet, to fuse RGB and pseudo-LiDAR information. 

\item To further enhance the information fusion network, we propose an information propagation path search approach (IPPS) to automatically tune and optimize our manually designed PLIFNet. With RGB images being the only input, our optimized network outperforms the current state-of-the-art RGB and LiDAR-based methods.

\item We propose a modality distillation strategy to free our PLIFNet from additional computational costs due to the generation of pseudo-LiDAR, making the pipeline of using PLIFNet more concise and efficient.

\end{itemize}

\section{Related Work}
\label{sec2}
Our model is related to road detection, neural architecture search (NAS), and knowledge distillation. Thus, we focus on discussing related works in these three areas.

\subsection{Road Detection}
Traditional road detection methods~\cite{he2004color, alvarez2014combining, alvarez2010road, lombardi2005switching} are usually based on edges, colors, and textures.
Alvarez and Lopez~\cite{alvarez2010road} introduced illuminant-invariant features which can be combined with a road class-likelihood classifier to obtain the road detection results.
He et~al.~\cite{he2004color} presented an approach in which boundaries are estimated on intensity images and road areas are detected using the full color images. Alvarez et~al.~\cite{alvarez2014combining} proposed a method to estimate road priors and combine different contextual cues, including horizon lines, vanishing points, lane markings, 3D scene layout, and road geometry, to obtain detection results.
Jeong et~al.~\cite{jeong2005efficient} proposed a classification method for road detection in which images are divided into sub-images, and a K-means-based classifier is used to perform classification for sub-images. 
Kumar et~al.~\cite{satzoda2015enhancing} designed a method to incorporate contextual cues to find lanes between lane curves and improve the computational efficiency without compromising accuracy. These traditional methods are proposed to solve
similar problems as ours, i.e., detecting drivable regions for autonomous vehicles. However, traditional methods are generally less effective than
current learning-based methods in terms of accuracy, as shown in the recent literature~\cite{sun2019reverse, chen2019progressive, fan2020sne}, 
and thus, we mainly focus on learning-based methods.

With the development of convolutional neural networks (CNNs), learning-based approaches are widely used in the field of semantic segmentation~\cite{long2015fully, chen2017deeplab, chen2018encoder, zhao2018icnet, meng2019weakly, ji2020encoder, lv2020weakly
}. 
Based on these semantic segmentation networks, a large number of CNN-based road detection methods~\cite{chen2017rbnet, sun2019reverse, chen2019progressive, bucher2019zero, fan2020sne} have been proposed for the road detection task. There is a strong correlation between semantic segmentation and road detection, as both of them can be regarded as 
pixel-wise labeling tasks. However,
compared with semantic segmentation, road detection only focuses on road regions and has higher requirements for prediction performance.

Qi et~al.~\cite{wang2017embedding} introduced a learning-based method to fuse RGB images, contours, and location priors in a network to predict road detection results.
Chen and Chen~\cite{chen2017rbnet} proposed a network in which they formulate the road detection and road boundary detection problem into a unified Bayesian network model to detect roads and road boundaries.
Sun et~al.~\cite{sun2019reverse} proposed a residual learning-based network with a residual refinement module composed of reverse and boundary attention units to perform road detection. Chen et~al.~\cite{chen2019progressive} presented an approach to transform LiDAR data space to RGB data space  
to combine two types of information together to obtain road detection results. Compared with the methods that only use RGB images, recent LiDAR-based methods~\cite{chen2019progressive, fan2020sne} have shown obvious accuracy advantages, which inspired us to propose our pseudo-LiDAR based framework.

\subsection{Neural Architecture Search}

 Neural architecture search (NAS) aims to discover high-performance neural architectures, enabling researchers to discard the tedious and heuristic manual neural architecture design process. A widespread phenomenon is that architectures built by NAS algorithms show better performance than manually designed architectures. Evolutionary algorithms (EAs) are used by early methods for optimizing neural architectures and parameters. The best architecture can be obtained by iteratively mutating a set of candidate architectures~\cite{liu2017hierarchical}.
An alternative to EAs is using reinforcement learning (RL) methods, e.g., policy gradients~\cite{zoph2018learning} and Q-learning~\cite{zhong2018practical}, to train a recurrent neural network as a meta-controller to generate potential architectures. 
Recently, a large number of speed-up methods~\cite{zhang2018graph, elsken2018efficient, pham2018efficient, liu2018darts} have been proposed to accelerate search speed, and NAS algorithms have been introduced into various tasks. For image classification, based on the continuous relaxation of the architecture representation, Liu et~al.~\cite{liu2018darts} proposed DARTS, in which gradient information is used to obtain the most useful path. For semantic segmentation, Auto-DeepLab~\cite{liu2019auto} expanded the search space of DARTS by introducing multiple paths with different widths and resolutions. For image restoration, Suganuma et~al. proposed E-CAE~\cite{suganuma2018exploiting}, where EA is adopted to search the architecture of convolutional auto-encoders for image inpainting and denoising.

In this work, we propose an NAS algorithm to search for information propagation paths to optimize the road detection network. The proposed IPPS is mostly related to DARTS~\cite{liu2018darts}. Both of them are gradient-based NAS algorithms, but the search space and target are different. DARTS is proposed to search for a classification network.
The IPPS is proposed to further optimize the road detection network by finding valuable paths. The search space of DARTS is on all candidate operations, while the search space of our IPPS is on all candidate information propagation paths. 
To our knowledge, our IPPS is the first NAS work that is proposed to fine-tune the information propagation paths in a manually designed network.

\subsection{Knowledge Distillation}

In the work of~\cite{hinton2015distilling}, knowledge distillation is introduced by distilling the knowledge in an ensemble of models into a single model to obtain results on MNIST. 
In recent years, a large number of knowledge distillation 
methods~\cite{chen2017learning, yim2017gift, yu2017visual, lee2018self, wang2018kdgan, cho2019efficacy,tung2019similarity, jin2019knowledge,peng2019correlation} 
have been proposed. Knowledge distillation are widely used in different types of tasks, such as object detection~\cite{chen2017learning, hao2019end}, image classification~\cite{park2019relational, liu2018multi}, and semantic segmentation~\cite{liu2020structured}. Chen et~al.~\cite{chen2017learning} proposed an end-to-end approach to learn compact multi-class object detection models through knowledge distillation. Park et~al.~\cite{park2019relational} presented a relational knowledge distillation method and showed its application in image classification and metric learning. Tang et~al.~\cite{tang2018ranking} proposed ranking distillation, which is the first knowledge distillation approach used in recommender systems. Pan et~al.~\cite{pan2019novel} designed an enhanced collaborative denoising auto-encoder (ECAE) model for recommender systems, in which a knowledge distillation is used to learn useful knowledge and reduce noise. Garcia et~al.~\cite{garcia2018modality} proposed a modality distillation approach to teach a hallucination network to mimic the depth stream in a multiple stream action recognition network. 
These distillation approaches can make networks to maintain their effectiveness on platforms with limited computing or input resources, which has significantly expanded their application scope.

Inspired by these distillation works, we design a modality distillation approach to reduce the number of parameters and computational costs of our framework. {We notice that a similar modality distillation idea has been used in CMT-CNN}~\cite{xu2017learning}. {However, CMT-CNN is proposed for pedestrian detection while our method is proposed for road detection. 
Different network architectures are also independently used in the two methods to meet different task requirements. During inference, CMT-CNN does not need to input thermal maps, but it relies on a region reconstruction network (RRN), while our method does not have such a requirement.}

 
\section{Our Method}
In this section, we elaborate on our pseudo-LiDAR based road detection framework, which is inspired by {previous pseudo-LiDAR based works}~\cite{wang2019pseudo, weng2019monocular, you2019pseudo} and LiDAR-based works~\cite{chen2019progressive, fan2020sne}.
To begin with, we introduce our framework to fuse RGB and pseudo-LiDAR information in Section \ref{sec:ddif}. Then we explain how to automatically tune and optimize the manually designed road detection network using the proposed IPPS in Section \ref{sec:ipps}. Finally, in Section \ref{MDsection} we present our modality distillation strategy, which is used to free our network from additional computation costs due to the use of depth estimation networks.

\begin{figure}[t]
\begin{center}
\includegraphics[width=3.49 in]{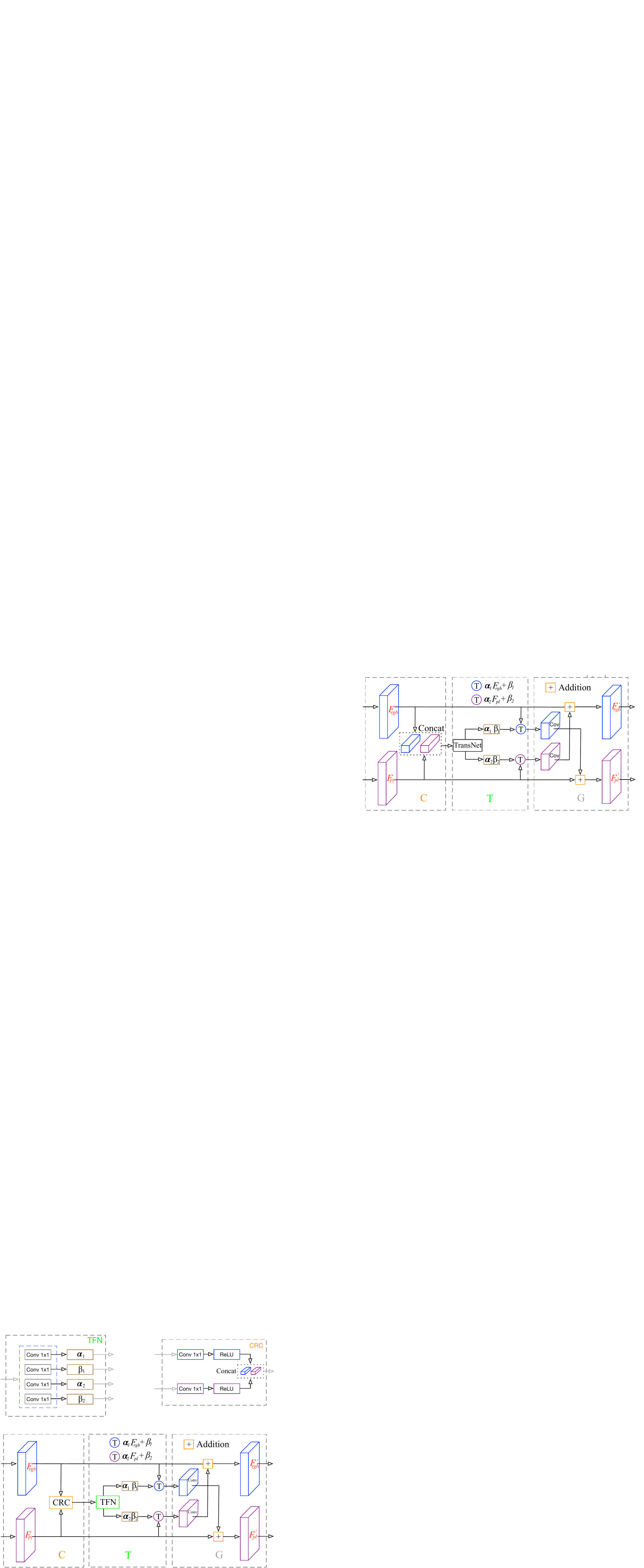}
\end{center}
  \caption{The detailed information fusion module CTG. The CTG module consists of three parts, i.e., concatenation (C), transformation (T), and generation (G). Firstly, RGB and PL features are concatenated together after the 1$\times$1 convolution and ReLU operations (top right). Then, the concatenated features are fed into a `TFN' module (top left) to obtain the $\alpha$ and $\beta$ parameters for fusion. Finally, the updated features are generated based on $\alpha$ and $\beta$.}
\label{fig:fusion_component}
\end{figure}

\subsection{Pseudo-LiDAR Based Road Detection Framework} 
\label{sec:ddif}  

The proposed pseudo-LiDAR based road detection framework is shown in Fig.~\ref{fig:network_overview}. It consists of three sub-components: pre-processing, information fusion, and post-processing. First, pseudo-LiDAR is generated in the pre-processing step. Then, the RGB and PL information are fused in the information fusion component. Finally, the RGB and PL features are combined together in the post-processing component to output the final detection results.

\noindent\textbf{Pre-processsing.} 
Current state-of-the-art methods~\cite{chen2019progressive, fan2020sne} usually need to use LiDAR data to improve their performance. However, as LiDAR data must be collected using specific devices, these methods are no longer applicable when only cameras are available. To eliminate the requirement on the use of LiDAR data, we introduce PL in our framework. For each RGB image, we first use a depth estimation network to generate its depth. Then we transform the learned depth into PL. In our framework, the BTS~\cite{lee2019big} network is used for depth estimation.
{Before training the road detection network, the depth network is trained independently using ground truth depth. When training the road detection network, the depth network is only used to generate PL and will not be trained again.
Note that the BTS network can be replaced by any other monocular depth estimation networks}~\cite{cao2017estimating, yan2017single, godard2017unsupervised, mohaghegh2018aggregation, zheng2018net, wang2019structure}.

Similar to the LiDAR-based work~\cite{chen2019progressive}, we transform all the depth values to their attitude space to generate PL.
For each pixel $(u, v)$ in a depth image, its attitude value $H_{(u , v)}$ in the camera coordinate system can be derived as follows:
\begin{equation}
H_{(u , v)} = \frac{(v - c_v)}{f_v} \times z \quad {\rm with }  \quad z = D_{(u, v)}, 
\end{equation}
where $D_{(u, v)}$ represents the depth value of pixel $(u, v)$, and $f_v$ and $c_v$  are camera focal length and the $v$ component of the principal point position $(c_u, c_v)$. 

After transforming all the depth values to attitude values, the $PL(u , v)$ which describes the pseudo-LiDAR value at coordinate position $(u, v)$ is defined as:

\begin{equation}
\begin{split}
PL_{(u , v)} =  \frac{1}{N}\sum_{N_u, N_v}\sqrt{\left(\frac{\delta_h}{N_u - u}\right)^2 + \left(\frac{\delta_h}{N_v - v}\right)^2} \\
{\rm with } \quad \delta_h = H_{(N_u, N_v)} - H_{(u, v)},
\end{split}
\end{equation}
where $(N_u, N_v)$ represents neighbourhood positions of $(u, v)$ and $N$ is the total number of neighbourhood positions. The $(N_u, N_v)$ is set as positions within a 7$\times$7 window centered at $(u, v)$.

\noindent\textbf{Information Fusion.} 
We use ResNet~\cite{he2016deep} as the backbone to build a two-branch fusion structure. The main layers in the ResNet are denoted as stage 1 to stage 4. RGB features and PL features are fused to exchange their information after each stage. The detailed process to fuse and update RGB and PL features after each stage is shown in Fig.~\ref{fig:fusion_component}. {The proposed information fusion module CTG consists of three sub-components, i.e., concatenation (C), transformation (T), and generation (G). In component C, a Conv-ReLU-Concat (CRC) module is constructed to concatenate features from two branches. Due to the difference between RGB information and PL information, the features extracted from RGB images are inconsistent with the features extracted from PL. To solve this feature space inconsistency problem, the RGB and PL features are transformed to each other's data space in the information fusion process. The data space transformation is treated as a linear transformation, and
the component T is used for generating transformation parameters $\alpha_{1}$, $\beta_{1}$, $\alpha_{2}$, and $\beta_{2}$. In component T, a transformation network (TFN) inputs the merged features and outputs the transformation parameters. Finally, the fused features are generated in component G as follows:} 
\begin{equation}
F_{rgb}' = F_{rgb} + (\alpha_{1}F_{pl} + \beta_{1}),
\label{transformation_eq1}
\end{equation}
\begin{equation}
F_{pl}' =  F_{pl} + (\alpha_{2}F_{rgb} + \beta_{2}),
\label{transformation_eq2}
\end{equation}
where $F_{rgb}$ and $F_{rgb}'$ represent the RGB features before and after information fusion respectively, while $F_{pl}$ and $F_{pl}'$ represent the PL features before and after information fusion respectively.

\noindent\textbf{Post-processing.}
After four stages of information exchange and fusion, features from the RGB branch and the PL branch are combined in the post-processing component to predict the final road detection results.
In the post-processing component, the RGB and PL features are sent to the pyramid pooling module (PPM)~\cite{zhao2017pyramid} to fuse features from different pyramid scales. For the details of PPM operation, we refer to~\cite{zhao2017pyramid}. Then, the RGB and PL information are combined together using a concatenation operation. Finally, classification convolutions (Conv)~\cite{zhao2017pyramid} are performed to predict results.

\begin{figure}[t]
\begin{center}
\includegraphics[width=3.4in]{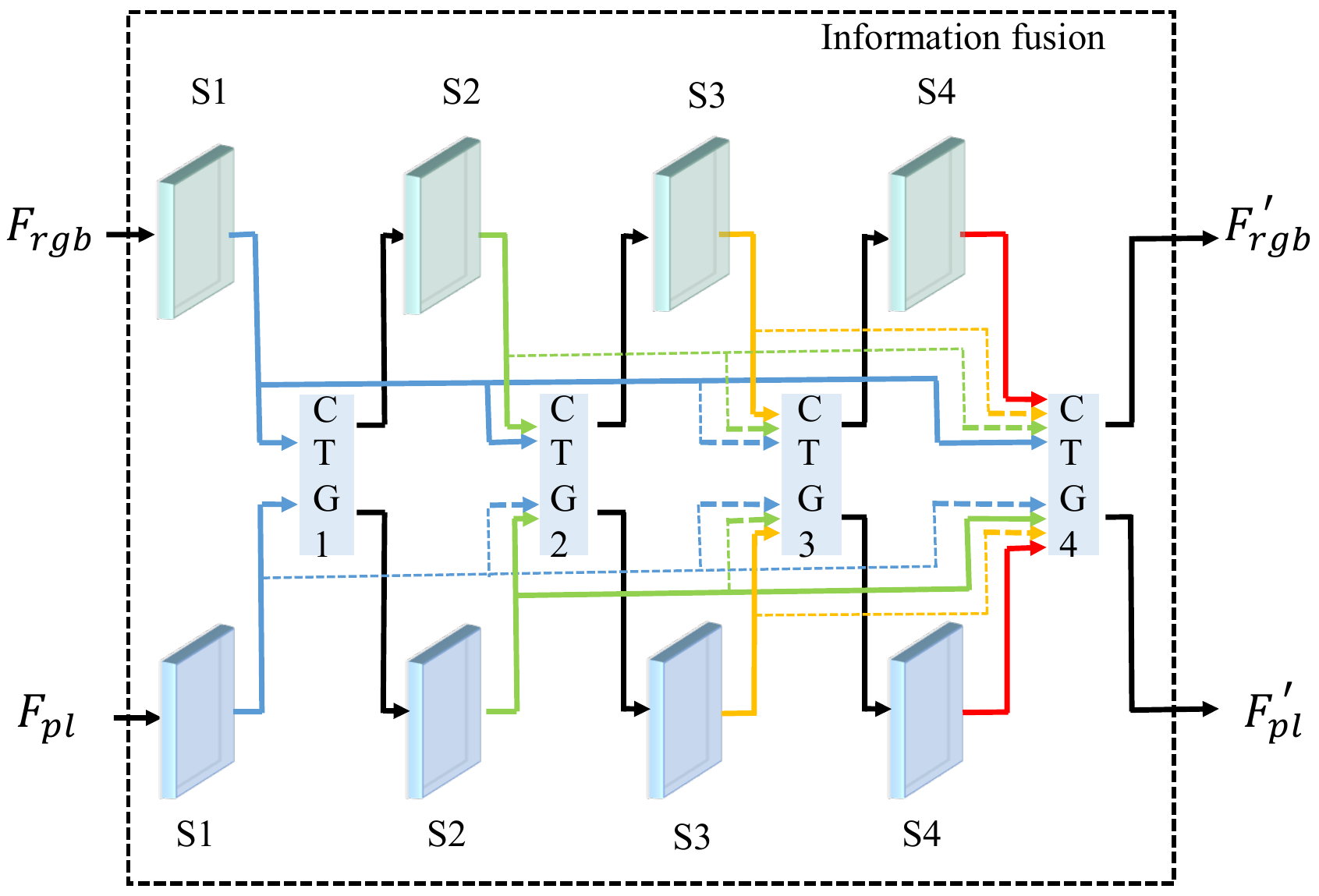}
\end{center}
\caption{Searching for information propagation paths in the information fusion component. Arrows with colors of blue, green, yellow, and red denote features from stage 1 to 4 respectively. Arrows in bold represent the paths searched by our proposed IPPS algorithm.}
\label{fig: IPPS}
\end{figure}

\subsection{Information Propagation Path Search}
\label{sec:ipps}  
In deep learning models, the features from shallow layers contain rich detailed information and the features from deep layers reveal rich semantic information. 
Fusing shallow features and deep features is beneficial for keeping output details. Currently, most of the dense prediction networks need to combine features from different layers to improve the prediction accuracy. As road detection needs to label each pixel in an image, it can also be regarded as a dense prediction task.
Therefore, we can further improve the performance of the proposed road detection network by adding information propagation paths between shallow layers and deep layers. 

Adding information propagation paths can combine information from shallow layers and deep layers, but it causes a problem as to how to find the paths. Furthermore, adding too many paths will introduce unnecessary parameters and make the network overloaded.  On the other hand, adding too few paths cannot make full use of the shallow and deep features. 

To solve this problem, we propose an information propagation path search (IPPS) algorithm to search for the most valuable paths. As shown in Fig. \ref{fig: IPPS}, we first build an overloaded information fusion module by adding all the possible information propagation paths in the original information fusion module as presented in Section~\ref{sec:ddif}. Specifically, in the original information fusion module, each CTG module only takes features from its current stage as input, while in the overloaded information fusion module, each CTG module takes not only the features from its current stage but also the features from all the previous stages as input. We denote the CTG module in stage i as ${CTG}_{i}$. In the original CTG module, the output of ${CTG}_{i}$ is calculated as:

\begin{equation}
{h}_{i}^{rgb}, {h}_{i}^{pl} = {CTG}_{i}({s}_{i}^{rgb}, {s}_{i}^{pl}),
\end{equation}
where ${s}_{i}^{rgb}$ and ${s}_{i}^{pl}$ represent the output of the RGB channel and the PL channel of the $i$th stage. ${h}_{i}^{rgb}$ and ${h}_{i}^{pl}$ are the outputs 
of ${CTG}_{i}$.
During  the  searching  process, the output of ${CTG}_{i}$ is:
\begin{equation}
\begin{split}
{}& {h}_{i}^{rgb}, {h}_{i}^{pl} = {CTG}_{i}({o}_{i}^{rgb}, {o}_{i}^{pl}), \\
{}& {o}_{i}^{c} = \sum _{j=1}^{i }{{p}_{c,i}^{j\rightarrow i}{T}_{c, i}^{j\rightarrow i}({s}_{j}^{c})},
\end{split}
\end{equation}
where $p$ is a set of weights of the information propagation paths. ${T}_{c,i}^{j\rightarrow i}$ denotes the transfer function between the $j$th stage and the $i$th stage in channel $c$. ${T}_{c,i}^{j\rightarrow i}$ consists of a single convolution layer, which is used to make the dimension and spatial resolution of ${s}_{j}^{c}$ consistent with that of ${s}_{i}^{c}$. When $j$ is equal to $i$, it is an identity mapping. The purpose of searching information propagation paths is to learn $p$, which is updated via gradient descent during the searching process. The optimization function of the searching process can be expressed as:
\begin{equation}
\mathop{\arg\min}_{p, \theta} Segloss(F(x,y|(p, \theta))),
\end{equation}
where $Seg loss$ is the cross entropy loss for road detection, $F()$ denotes the road detection network, $x$ and $y$ are the input image and the corresponding ground truth label respectively, and $\theta$ represents the kernels of convolution layers of the network. During the searching process, $p$ and $\theta$ are updated alternately. After optimization, for each stage, we keep the path which has the maximum weight. If the maximum weight is negative, we will discard all paths from shallow layers to the current stage.

\begin{figure}[t]
\begin{center}
\includegraphics[width=3.49in]{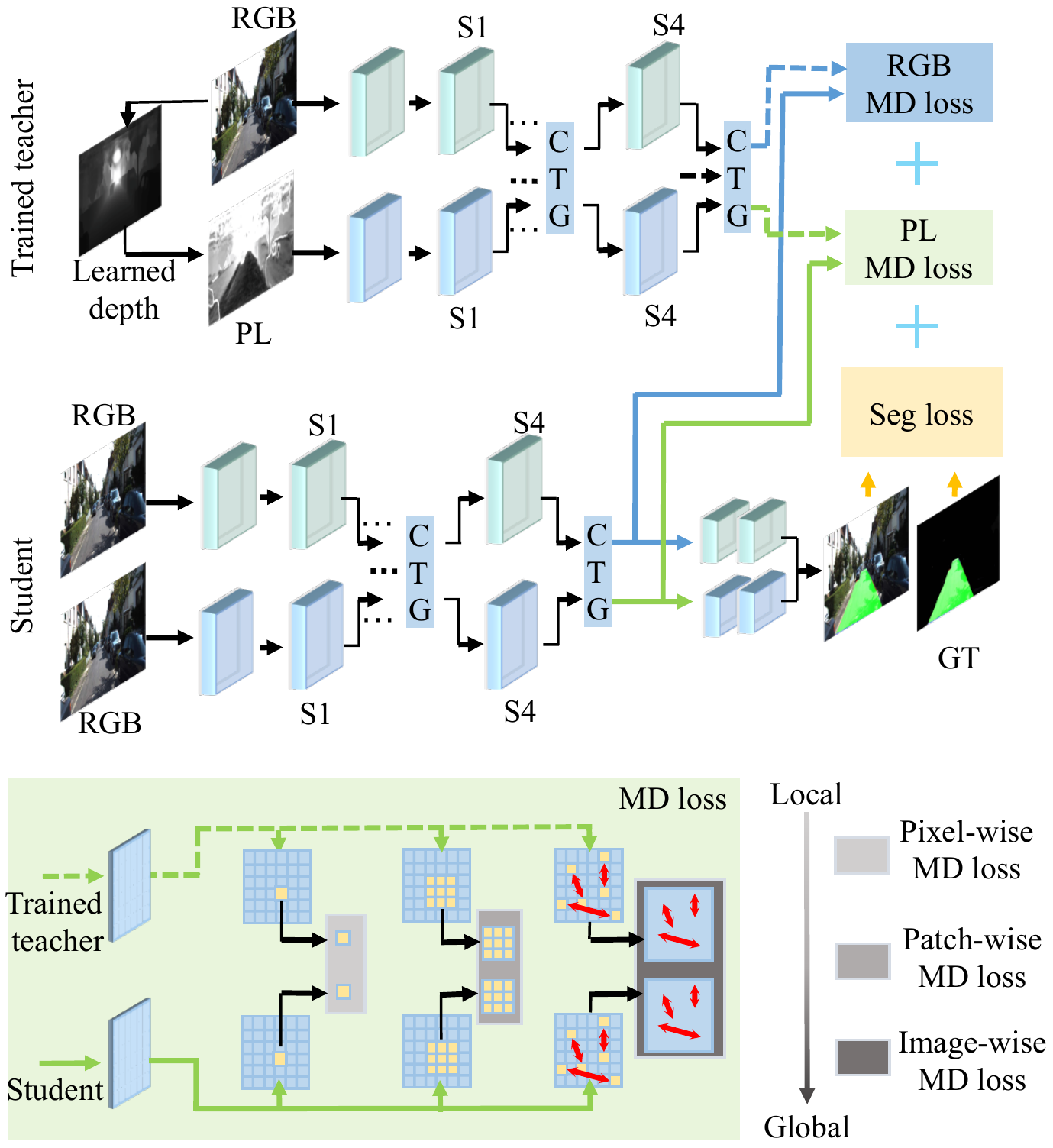}
\end{center}
  \caption{Modality distillation (MD) framework and MD loss. During inference, the student network only needs to take RGB images as inputs. From local to global, our proposed MD loss consists of three parts, i.e., pixel-wise loss, patch-wise loss, and image-wise loss.}
\label{fig: MD}
\end{figure}

\subsection{Modality Distillation}
\label{MDsection}
By using the architecture optimization of IPPS, our framework can achieve state-of-the-art performance. However, generating PL relies on depth  estimation  networks and introduces  additional computation costs. To reduce computational costs and make our framework more concise during inference, we propose a modality distillation (MD) method. 

As shown at the top of Fig.~\ref{fig: MD}, our MD framework consists of two main parts, the trained teacher part and the student part. Two channels of the trained teacher network take RGB and PL information as inputs. In contrast, both of the two channels of the student network take RGB images as inputs. During training, the student network is trained to learn the features from the teacher network by using different losses. The optimization function is defined as:
\begin{equation}
\begin{aligned}
&{L}_{MD} = \lambda{l}_{MD, RGB} + \lambda{l}_{MD, PL} + {l}_{seg},\\
&{l}_{MD, RGB} = {l}_{pi}({h}_{t}^{1}, {h}_{s}^{1}) + {l}_{pa}({h}_{t}^{1}, {h}_{s}^{1}) + {l}_{im}({h}_{t}^{1}, {h}_{s}^{1}),\\
&{l}_{MD, PL} = {l}_{pi}({h}_{t}^{2}, {h}_{s}^{2}) + {l}_{pa}({h}_{t}^{2}, {h}_{s}^{2}) + {l}_{im}({h}_{t}^{2}, {h}_{s}^{2}),    
\end{aligned}
\end{equation}
where ${h}_{t}^{1}$ and ${h}_{s}^{1}$ represent the outputs of channel 1 of the last CTG modules of the trained teacher and student networks respectively. ${h}_{t}^{2}$ and ${h}_{s}^{2}$ denote the outputs of channel 2 of the last CTG modules of the trained teacher and student networks respectively. ${l}_{seg}$ is the segmentation loss, which is defined as the cross entropy loss between the prediction and ground truth. ${l}_{MD, RGB}$ and ${l}_{MD, PL}$ are the modality distillation losses of the RGB and PL channels respectively. From local to global, each MD loss consists of three items: pixel-wise MD loss (${l}_{pi}$), patch-wise MD loss (${l}_{pa}$), and image-wise MD loss (${l}_{im}$). 

\noindent\textbf{Pixel-wise MD loss.} This is the loss item that restricts the distance between features for pixel pairs. We use the squared difference to formulate this loss:
\begin{equation}
{l}_{pi}({h}_{t}, {h}_{s})=\frac{\sum_{i=1}^{{H}}{\sum_{j=1}^{W}{(f({h}_{t}^{ij})-f({h}_{s}^{ij}))}^{2}}}{H\times W}
\end{equation}
where $H$ and $W$ denote the spatial resolution of the feature map. $f()$ is the sigmoid function. ${h}_{t}^{ij}$ is the feature vector at location $(i,j)$ of the trained teacher network, and ${h}_{s}^{ij}$ is the feature vector at location $(i,j)$ of the student network.

\begin{figure*}[t]
\begin{center}
\includegraphics[width=1.0\textwidth]{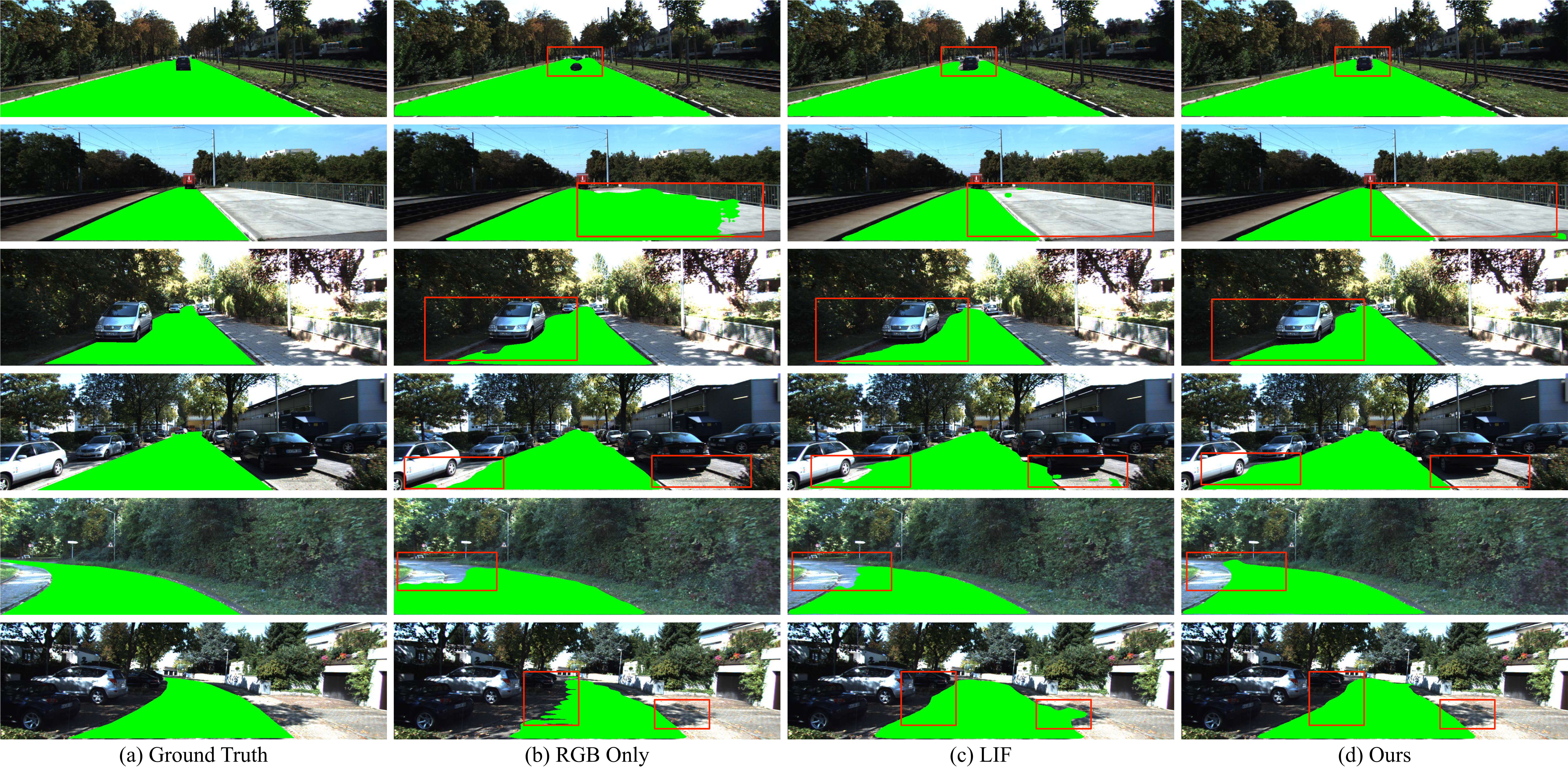}
\end{center}
  \caption{Visualization comparisons on the KITTI dataset. The comparison details are marked with red boxes. Three types of visualization results are presented, i.e., RGB only, LIF~\cite{chen2019progressive}, and our optimized PLIF.}
\label{visualcmp}
\end{figure*}

\begin{table}[t]
\footnotesize
\renewcommand\arraystretch{1.0}
\caption{Benefits of PL based information fusion. `NF' denotes no information fusion; `LIF' represents the information fusion approach used in~\cite{chen2019progressive}; `PLIF' indicates our proposed PL based information fusion. Results are based on 5-fold cross-validation on the KITTI dataset. Best scores are in bold.}
\begin{center}
{
\begin{tabular}{l|c|c|c|c}
\hline
Method&Input& MaxF (\%) & AP (\%) & PRE (\%) \\
\hline
NF & RGB & 94.88 & 92.19 & 94.97  \\
NF & Depth & 95.07 & 92.19 & 95.44  \\
NF & PL & 95.30 & 91.95 & 95.44  \\
\hline
LIF & RGB + PL & 95.44 & 92.26 & 95.60  \\
PLIF & RGB + PL & \textbf{96.06} & \textbf{92.30} & \textbf{95.95}\\
\hline
\end{tabular}
}
\end{center}
\label{Table: KITTICross}
\end{table}

\begin{table}[t]
\footnotesize
\renewcommand\arraystretch{1.0}
\caption{Comparison between using IPPS paths and using all propagation paths. }
\begin{center}
{
\begin{tabular}{l|c|c|c|c}
\hline
Method & MaxF (\%) & AP (\%) & PRE (\%) & REC (\%)\\
\hline
All Path & 95.25 & 92.09 & 95.24 &95.26 \\
\hline
IPPS Path & \textbf{96.23} & \textbf{92.22} & \textbf{96.14} &\textbf{96.32} \\
\hline
\end{tabular}
}
\end{center}
\label{Table: KITTINAS}
\end{table}

\begin{table}[t]
\footnotesize
\renewcommand\arraystretch{1.0}
\caption{Performance comparison between using ground truth (GT) depth and learned depth (LD). Results are based on the R2D dataset and best scores are in bold.}
\begin{center}
{
\begin{tabular}{c|c|c|c|c}
\hline
Input & ACC (\%) & F-Score (\%) & PRE (\%) & REC (\%) \\
\hline
LD & 98.72  & 98.42   & 97.70  & 99.15 \\
\hline
GT & \textbf{99.30}& \textbf{99.13} &  \textbf{98.90}&  \textbf{99.36} \\
\hline
\end{tabular}
}
\end{center}
\label{Table: R2D_Depth}
\end{table}

\begin{table}[t]
\footnotesize
\renewcommand\arraystretch{1.0}
\caption{Effectiveness of modality distillation. Three components are evaluated, i.e., PI (pixel-wise MD loss), PA (patch-wise MD loss), and IM (Image-wise MD loss). Results are based on 5-fold cross-validation on the KITTI dataset. Best scores are in bold.}
\begin{center}
{
\begin{tabular}{ccc|c|c|c|c}
\hline
PI & IM & PA & MaxF (\%) & AP (\%) & PRE (\%) & REC (\%) \\
\hline
- & - & - & 94.72 & 92.09 & 94.75  & 94.70  \\
 \checkmark & - & - & 95.10  & 92.19  & 95.19 & 95.01  \\
 \checkmark & \checkmark  & - & 95.23 & 92.17 & 95.21  &95.25\\
 \checkmark & \checkmark & \checkmark & \textbf{95.41} & \textbf{92.32} & \textbf{95.55} &\textbf{95.26} \\
 \hline
\end{tabular}
}
\end{center}
\label{Table: KITTIMod}
\end{table}

\noindent\textbf{Patch-wise MD loss.} This is the loss item that restricts the structure similarity of feature patches. This loss is calculated as:
\begin{equation}
{l}_{pa}({h}_{t}, {h}_{s})={\log}_{10}({\rm ssim}(
{h}_{t}, {h}_{s}
)^{-1}),   
\end{equation}
where $\rm ssim()$ is structural similarity~\cite{wang2004image}. The window size of $\rm ssim()$ is set to 11 in this paper.

\noindent\textbf{Image-wise MD loss.} This is the loss term that is used to maintain structure consistency of the whole feature map. The loss function is defined as:
\begin{equation}
{l}_{im}({h}_{t}, {h}_{s})=\frac{1}{n}\sum_{r=1}^{n} \frac{{\rm{cos}}^{-1}({\rm Sim}({h}_{t,r}^{\prime}, {h}_{s,r}^{\prime}))}{\pi},
\end{equation}
where $n$ is the number of random sampling, and $\rm Sim()$ means similarity, which is defined as:
\begin{equation}
\begin{aligned}
&{\rm Sim}({h}_{t,r}^{\prime}, {h}_{s,r}^{\prime})=\frac{{h}_{t,r}^{\prime}\cdot {h}_{s,r}^{\prime}}{||{h}_{t,r}^{\prime}||\ ||{h}_{s,r}^{\prime}||},\\
& {h}_{t,r}^{\prime}= {h}_{t}^{{i}_{r}{j}_{r}}-{h}_{t}^{{i}_{r}^{\prime}{j}_{r}^{\prime}}, \\
&{h}_{s,r}^{\prime}= {h}_{s}^{{i}_{r}{j}_{r}}-{h}_{s}^{{i}_{r}^{\prime}{j}_{r}^{\prime}},
\end{aligned}
\end{equation}
where $\cdot$ is inner product and $||\ ||$ represents the norm of a vector. $({i}_{r}, {j}_{r})$ and $({i}_{r}^{\prime}, {j}_{r}^{\prime})$ are two sets of coordinate positions which are randomly selected in the $r$th sampling.

\section{Experiments}
We perform comprehensive ablation studies and comparisons on the KITTI~\cite{fritsch2013new} and R2D~\cite{fan2020sne} datasets to investigate the effectiveness of our method. First, to investigate the effectiveness of each part in our method, we adapt 5-fold cross-validation to perform ablation studies on the KITTI dataset. Then, we compare our method with other state-of-the-art methods on both the KITTI and R2D datasets.

\begin{table}[t]
\footnotesize
\renewcommand\arraystretch{1.0}
\caption{{Using different pseudo-LiDAR representations. `PLXYZ' represents using the pseudo-LiDAR in 3D objection methods}~\cite{wang2019pseudo, weng2019monocular}. {`PLDY' represents the pseudo-LiDAR in our framework.} }
\begin{center}
{
\begin{tabular}{c|c|c|c|c}
\hline
Transformation & MaxF (\%) & AP (\%) & PRE (\%) & REC (\%)\\
\hline

PLXYZ & 95.62 & \textbf{92.25} & 95.60 &95.65 \\

\hline
PLDY & \textbf{96.23} & 92.22 & \textbf{96.14} &\textbf{96.32} \\
\hline
\end{tabular}
}
\end{center}
\label{Table: pseudoLiDARCmp}
\end{table}

\begin{table}[t]
\footnotesize
\renewcommand\arraystretch{1.0}
\caption{Benefits of using modality distillation. `DepthNet' and `RoadNet' denote the network to generate depth and our road detection network respectively. `Full' represents our full method and `MD' denotes the proposed modality distillation method.}
\begin{center}
{
\begin{tabular}{l|c|c|c|c}
\hline
Metric& DepthNet & RoadNet & Full & MD \\
\hline
Speed & 0.21 s& 0.25  s& 0.21 + 0.25 s& 0.25 s\\
\hline
Params & 112.8 M & 97.89 M & 112.8 + 97.89 M & 97.89 M \\
\hline
\end{tabular}
}
\end{center}
\label{MDSpeed}
\end{table}

\begin{figure}[t]
\begin{center}
\includegraphics[width=3.4 in]{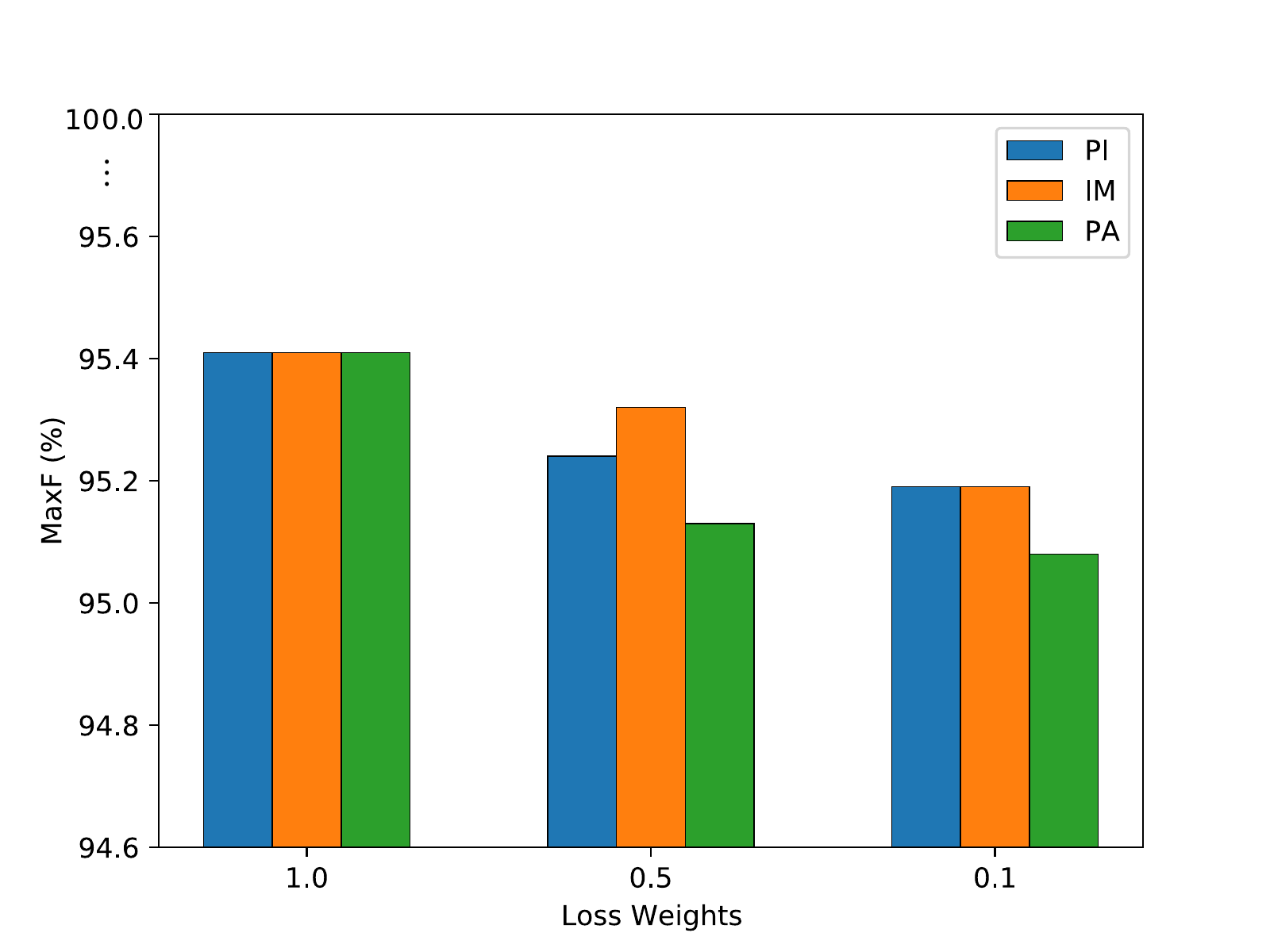}
\end{center}
\vspace{-0.2cm}
  \caption{{The performance of using different loss weights on the KITTI dataset. PI, PA, and IM represent pixel-wise loss, patch-wise loss, and image-wise loss respectively.
  To reduce the number of variables, each time only one loss weight variable is changed, while the other two are fixed to 1.0. For example, if the weight of PA is changed to 0.1, the weights of PI and IM are set to 1.0.}}
\label{loss_weight}
\end{figure}

\begin{figure}[t]
\begin{center}
\includegraphics[width=3.4 in]{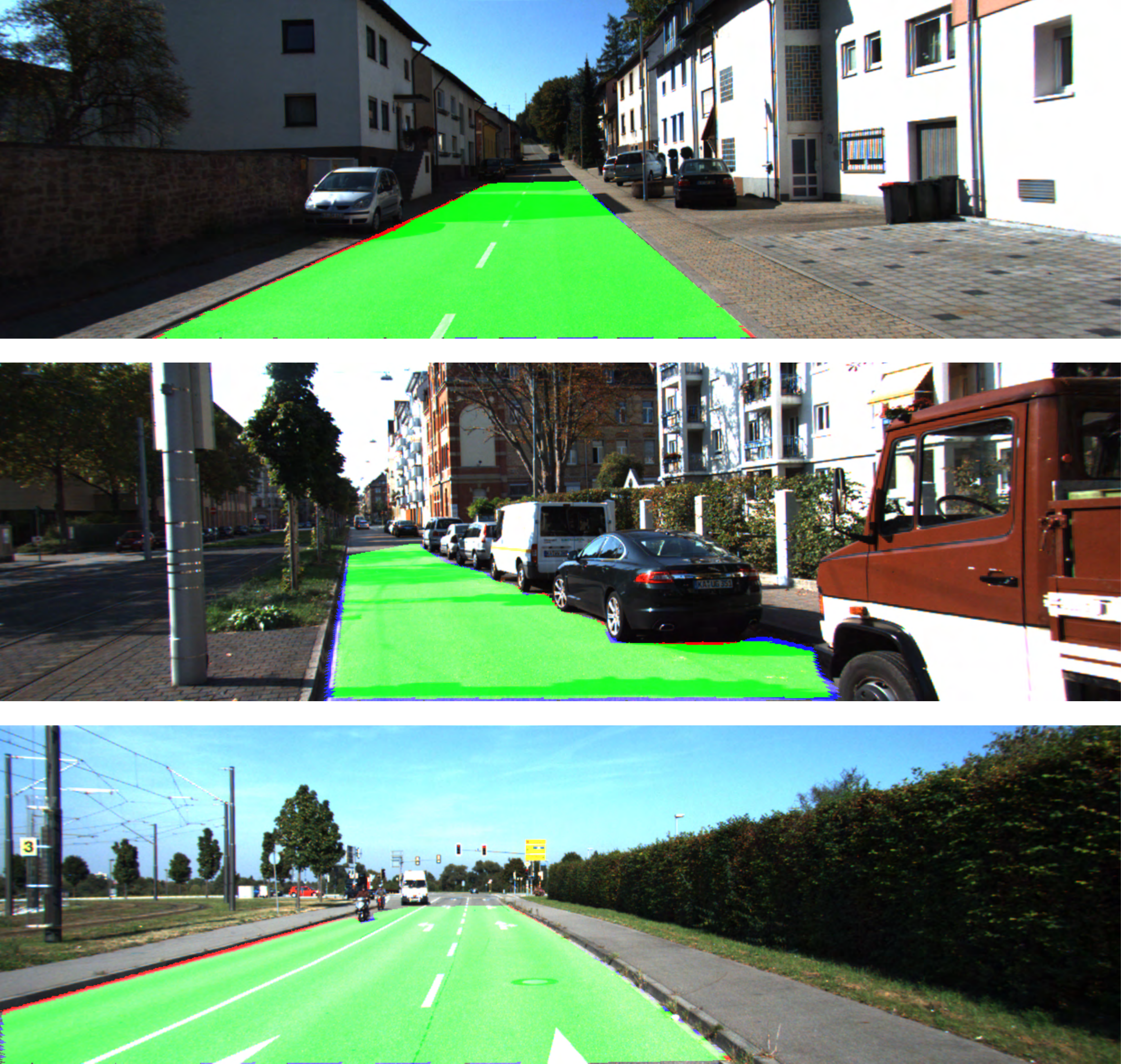}
\end{center}
  \caption{Road detection results on the KITTI benchmark~\cite{fritsch2013new}. Top: urban marked (UM), middle: urban unmarked (UU), bottom: urban multiple marked (UMM).}
\label{KITTI_bechmark_vis}
\end{figure}

\begin{figure*}[t]
\begin{center}
\includegraphics[width=6.9 in]{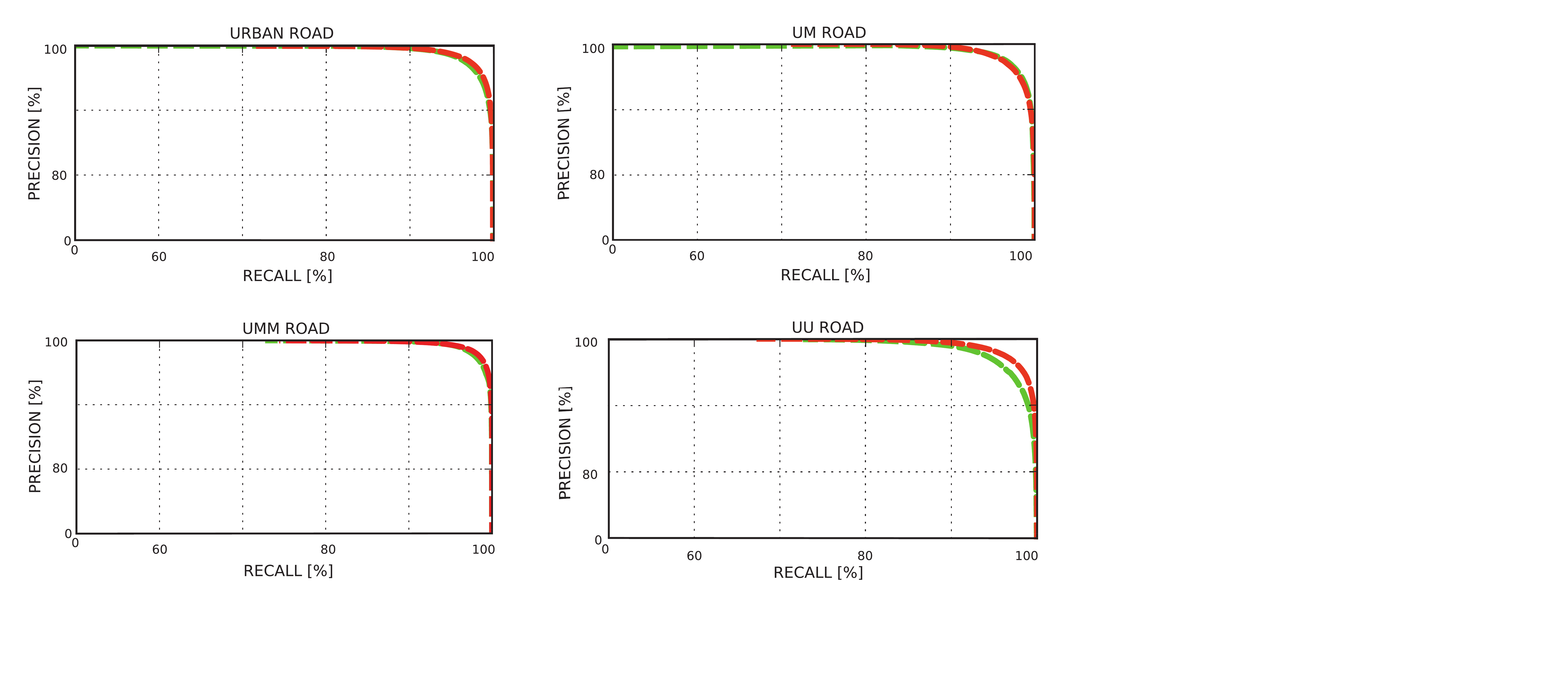}
\end{center}
  \caption{Precision and recall curves on the KITTI benchmark. Three categories are presented, i.e., urban  marked  roads (UM), urban  multiple  marked  roads (UMM) and urban unmarked roads (UU). Urban road is a combination of these three categories. Two methods are compared, i.e., PLARD (green lines) and ours (red lines).}
\label{pr_curves}
\end{figure*}
\begin{table*}[t]
\footnotesize
\renewcommand\arraystretch{1.0}
\caption{Comparison with the Top-10 published road detection methods on the KITTI benchmark (URBAN\_ROAD Ranking). Best scores are in bold. `Full' denotes using the proposed full information fusion network in which the learned depth is used to generate PL.}
\begin{center}
{
\begin{tabular}{l|c|cccccc|c}
\hline
Methods & Input & MaxF (\%) & AP (\%) & PRE (\%) & REC (\%) & FPR (\%) & FNR (\%) & Rank \\
\hline
RBNet~\cite{chen2017rbnet} & RGB & 94.97 & 91.49  & 94.94  & 95.01  & 2.79  & 4.99 & 11 \\
TVFNet~\cite{gu2019two} & RGB + LiDAR  & 95.34  & 90.26 & 95.73  & 94.94  & 2.33  & 5.06  & 10 \\
SSLGAN~\cite{han2018semisupervised} & RGB  & 95.53  & 90.35  & 95.84  & 95.24  & 2.28  & 4.76  & 9 \\
RGB36-Cotrain~\cite{caltagirone2019lidar} & RGB  & 95.55  & 93.71 & 95.68  & 95.42 & 2.37  & 4.58  & 8 \\
LC-CRF~\cite{gu2019road} & RGB + LiDAR  & 95.68  & 88.34 & 93.62 & \textbf{97.83} & 3.67 & \textbf{2.17 } & 7 \\
NIM-RTFNet~\cite{wang2020applying} & RGB + LiDAR & 96.02  & 94.01  & 96.43 & 95.62  & 1.95 & 4.38 & 6 \\
LidCamNet~\cite{caltagirone2019lidar2} & RGB + LiDAR & 96.03  & 93.93  & 96.23 & 95.83  & 2.07 & 4.17  & 5 \\
RBANet~\cite{sun2019reverse} & RGB & 96.30  & 89.72  & 95.14  & 97.50  & 2.75  & 2.50 & 4\\
SNE-RoadSeg~\cite{fan2020sne} & RGB + LiDAR & 96.75  & 94.07  & 96.90  & 96.61  & 1.70  & 3.39 & 3 \\
PLARD~\cite{chen2019progressive} & RGB + LiDAR & 97.03  & 94.03  & 97.19  & 96.88 & 1.54 & 3.12 & 2\\
\hline
Ours (Full) & RGB & \textbf{97.42} & \textbf{94.09} &\textbf{97.30} & 97.54  & \textbf{1.49} & 2.46 & 1 \\
\hline
\end{tabular}
}
\end{center}
\label{Table: KITTIBenchmark}
\end{table*}

\begin{figure*}[t]
\begin{center}
\includegraphics[width=7.1 in]{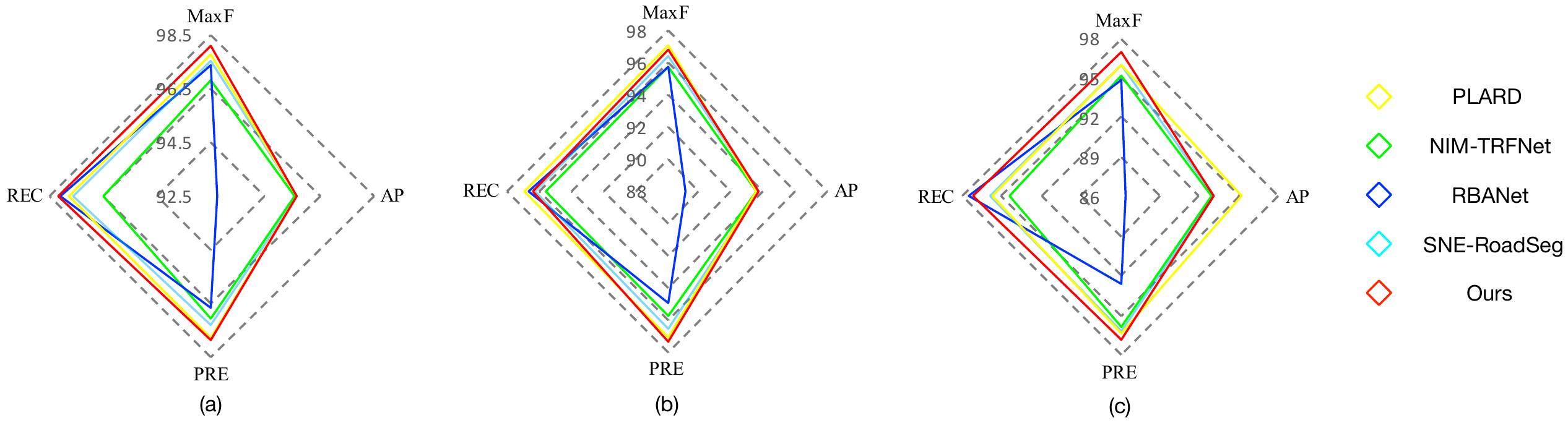}
\end{center}
  \caption{The performance of different categories on the testing set of the KITTI benchmark. (a) urban multiple marked roads (UMM); (b) urban marked roads (UM); (c) urban unmarked roads (UU).}
\label{umm_um_uu}
\end{figure*}

\begin{table*}
\caption{Comparisons of UU, UM, and UMM on the KITTI benchmark. Best `MaxF' scores are in bold. `Full' denotes using the proposed full information fusion network in which the learned depth is used to generate PL. }
\center
\begin{tabular}{l|c|c|c|c|c|c|c}
\hline 
\multirow{2}{*}{Method} & \multirow{2}{*}{Input}  & \multicolumn{2}{c|}{UU} & \multicolumn{2}{c|}{UM} & \multicolumn{2}{c}{UMM}\\
\cline{3-8} 
&  & MaxF (\%)& AP (\%) & MaxF (\%)& AP (\%)& MaxF (\%)& AP (\%)\\ 
\hline

RBNet~\cite{chen2017rbnet} & RGB & 93.21 & 89.18 & 94.77&91.42 & 96.06 &93.49 \\ 
TVFNet~\cite{gu2019two} & RGB + LiDAR  & 93.65 & 87.57 &94.96&89.17 & 	96.47&93.16  \\
SSLGAN~\cite{han2018semisupervised} & RGB  &94.40 &87.84 & 94.62&89.50 &96.72&92.99 \\
RGB36-Cotrain~\cite{caltagirone2019lidar} & RGB &94.53&92.54&94.55&93.12 & 96.75 & 95.39 \\
LC-CRF~\cite{gu2019road} & RGB + LiDAR  & 94.01 & 85.24 &94.91&86.41 & 97.08&92.06 \\
NIM-RTFNet~\cite{wang2020applying} & RGB + LiDAR &  	95.11 &92.94 &95.71 & 	93.56 &96.79 & 	95.61 \\
LidCamNet~\cite{caltagirone2019lidar2} & RGB + LiDAR & 94.54 & 	92.74&95.62&93.54 &97.08 &	95.51 \\
RBANet~\cite{sun2019reverse} & RGB &  	94.91 & 	86.35&95.78 &	89.14 &97.38 &92.67 \\
SNE-RoadSeg~\cite{fan2020sne} & RGB + LiDAR & 96.03 & 	93.03 &96.42 &93.67&97.47 & 	95.63\\
PLARD~\cite{chen2019progressive} & RGB + LiDAR & 95.95 & 	95.25 & 	\textbf{97.05} & 93.53 & 97.77&95.64\\
\hline
Ours (Full) & RGB & \textbf{96.93} & 	93.08 &96.87 &	{93.71}& \textbf{98.05} &95.63\\
\hline
\end{tabular}
\label{UUUM}
\end{table*}

\subsection{Datasets}
The KITTI road benchmark is one of the most popular road detection datasets. The KITTI road detection dataset contains three different categories of road scenes, i.e., urban unmarked roads (UM), urban multiple marked roads (UMM), and urban unmarked roads (UU). To avoid overfitting and carry out fair comparisons, the number of submissions is strictly restricted on the KITTI server. Thus, we investigate the effectiveness of each part of the proposed method by using 5-fold cross-validation on the KITTI training dataset, and only the final full method is evaluated on the KITTI benchmark server. 

The R2D dataset is a large-scale synthetic road detection dataset. It is collected in various weather conditions and contains 11,430 images with corresponding depth images and semantic labels. The authors of the R2D dataset split the whole dataset into three subsets, i.e., 6117 training images, 2624 validation images, and 2689 testing images. As the R2D dataset contains multiple scenarios in different environments, it can be used to evaluate the effectiveness of methods in different road scenes.

\subsection{Evaluation Metrics}
For the evaluation on the KITTI dataset, we use the standard evaluation metrics provided by the KITTI benchmark. The metrics consist of maximum F1-measure (MaxF), precision rate (PRE), average precision (AP), false positive rate (FPR), false negative rate (FNR), and recall rate (REC). For more details about these six metrics, we refer to~\cite{fritsch2013new}.

Similar to the evaluation on the KITTI dataset, five metrics are used by R2D to evaluate the performance. The metrics include accuracy (ACC), precision (PRE), recall (REC), F-score, and intersection over union (IoU). 
The details of the evaluation metrics on R2D are described in~\cite{fan2020sne}. For all the experiments on the R2D dataset, we follow the same evaluation approach as used in~\cite{fan2020sne}.

\subsection{Implementation Details}
We implement our network using PyTorch~\cite{paszke2019pytorch} and train it on NVIDIA V100 GPUs.
To obtain the high-quality depth to generate PL, the BTS~\cite{lee2019big} network is used for monocular depth estimation. The BTS network can be replaced by any other monocular depth estimation networks for PL generation. The experimental settings are similar in the ablation studies and the final evaluations. During training and testing, the input image size is resized to 384$\times$1280 for the KITTI dataset and 480$\times$640 for the R2D dataset. We use SGD to optimize all the parameters and the batch size is set to 4 on both the KITTI and R2D datasets. To better evaluate the effectiveness of each component of our network, models are trained from scratch without data augmentation in our ablation studies. For the final evaluations on the KITTI and R2D datasets, parameters from the model of~\cite{chen2019progressive} and our KITTI model are used as the initial weights for the training respectively. The initial learning rates are set to $1\times10^{-2}$ in our ablation studies and modality distillation. In addition, for the final benchmark evaluation, we use several data augmentation approaches to train our models, including random cropping and multi-scale training and testing.

\subsection{Ablation Studies}

\noindent\textbf{The Effectiveness of Information Fusion.} 
In this section, different approaches are compared to investigate the effectiveness of our information fusion approach, including NF, LIF, and our proposed PLIF. NF represents using only RGB, depth, or PL information as input. LIF means using both RGB and PL information, and only the information from the PL branch is fused into the RGB branch. PLIF denotes our proposed information fusion module, in which not only the PL information is fused into the RGB branch but also the RGB information is fused into the PL branch. Results of 5-fold cross-validation on the KITTI dataset are listed in Table~\ref{Table: KITTICross}. 
From the comparison results in Table~\ref{Table: KITTICross}, we can observe: 
(1) Using PL information can obtain a comparable performance with using RGB information. Thus, the PL information should be considered equally with the RGB information. Compared with using depth information as input, taking the PL information derived from depth information achieves better performance; (2) By considering PL and RGB information equally and making better use of all the information, our proposed PLIF achieves better performance than the LIF proposed in~\cite{chen2019progressive}. Using our PLIF approach increases the MaxF by 0.62\% compared with using the LIF approach.

\noindent\textbf{Benefits of Using IPPS.} We conduct experiments on the KITTI dataset to evaluate the effectiveness of the proposed IPPS method. The comparison results are shown in Table~\ref{Table: KITTINAS}. For the MaxF metric, we can observe that using all the information propagation paths only gets 95.25\% while using the IPPS path obtains 96.23\%. The main reason for this is that not all the paths are valuable for information fusion. The proposed IPPS method enables the network to select useful information propagation paths from all the propagation paths to obtain better accuracy. 

\noindent\textbf{Visualization Comparisons.}
In addition to the comparisons using performance metrics, visualization comparisons are also used to investigate the effectiveness of our method. For a better visualization comparison, some visualization results are highlighted in  Fig.~\ref{visualcmp}.
As can be seen from Fig.~\ref{visualcmp}, compared with using RGB information only or the LiDAR-based information fusion (LIF) network~\cite{chen2019progressive}, our proposed PLIF can obtain better prediction results. For example, in some boundary and shadow areas, our prediction results are closer to the ground truth. This is mainly because our PLIF network can make more effective use of RGB and pseudo-LiDAR information.


\noindent\textbf{Effects of Using Learned Depth.} As we use a depth estimation network to obtain learned depth to build our pseudo-LiDAR based framework, we perform an analysis to investigate the difference between using ground truth depth and learned depth. The analysis is performed on the R2D dataset, as the R2D datastet is a synthetic dataset and can provide enough reliable dense depth. We can observe from Table~\ref{Table: R2D_Depth} that using ground truth depth to replace learned depth can obtain better performance, but the performance cannot be significantly improved. There are two reasons for this situation. The first reason is that it is difficult to obtain further significant improvements, as our network has achieved very high performance. Another reason is that RGB information and pseudo-LiDAR information work together in our framework, which can make the network robust to slight depth errors.

\noindent\textbf{The Effectiveness of Modality Distillation.}
We perform comprehensive experiments to illustrate the effectiveness of our modality distillation approach on the KITTI dataset. 
The comparison results are shown in Table~\ref{Table: KITTIMod}. As can be seen from  Table~\ref{Table: KITTIMod}, from local to global, with the modality distillation loss items being added one by one, the accuracy of modality distillation shows steady growth. We can observe that by using all three proposed MD loss items, our modality distillation strategy increases MaxF from 94.72\% to 95.41\%. 

\begin{table*}[t]
\footnotesize
\renewcommand\arraystretch{1.0}
\caption{The modality distillation results and the comparison with other methods on the KITTI benchmark. Only the methods without using any refinement operation are compared. `MD' represents using the proposed modality distillation approach (without using the learned depth to generate PL in testing).}
\begin{center}
{
\begin{tabular}{l|c|c|c|c|c|c}
\hline
Methods & MaxF (\%) & AP (\%) & PRE (\%) & REC (\%) & FPR (\%) & FNR (\%) \\
\hline
RBNet~\cite{chen2017rbnet}  & 94.97 & 91.49  & 94.94  & 95.01  & 2.79  & 4.99\\
SSLGAN~\cite{han2018semisupervised}  & 95.53  & 90.35  & 95.84  & 95.24  & 2.28  & 4.76\\
RGB36-Cotrain~\cite{caltagirone2019lidar}  & 95.55  & 93.71 & 95.68  & 95.42 & 2.37  & 4.58\\
\hline
Ours (MD) & \textbf{95.79} & \textbf{93.95} & \textbf{96.12} & \textbf{95.46} & \textbf{2.12} & \textbf{4.54} \\
\hline
\end{tabular}
}
\end{center}
\label{KITTIBenchmarkMod}
\end{table*}

\begin{table*}[!htb]
\footnotesize
\renewcommand\arraystretch{1.0}
\caption{Performance comparison on the R2D dataset. The results of SNE-RoadSeg are from~\cite{fan2020sne}. `Full' represents our full method (leaned depth is used to generate the PL). `MD' represents using the proposed modality distillation approach (without using the learned depth to generate the PL during inference).}
\begin{center}
{
\begin{tabular}{l|c|c|c|c|c|c|c}
\hline
Methods & Backbone & Input & ACC (\%) & PRE (\%) & REC (\%) & F-Score (\%) & IoU (\%) \\
\hline
SNE-RoadSeg~\cite{fan2020sne}& ResNet-101 & RGBD  & 98.0 & 98.2 & 97.1 & 97.6 & 95.4 \\
SNE-RoadSeg~\cite{fan2020sne} & ResNet-152 & RGBD & 98.6 &  99.1 &  97.6 & 98.3 & 96.7  \\
\hline
Ours (MD) & ResNet-101& RGB & 99.1  & 99.2  & 98.5 & 98.9  & 97.8\\
Ours (Full) & ResNet-101& RGB & \textbf{99.5} & \textbf{99.5} & \textbf{99.3} & \textbf{99.4} & \textbf{98.8} \\
\hline
\end{tabular}
}
\end{center}
\label{Table: R2D}
\end{table*}

\begin{figure*}[t]
\begin{center}
\includegraphics[width=1.0\textwidth]{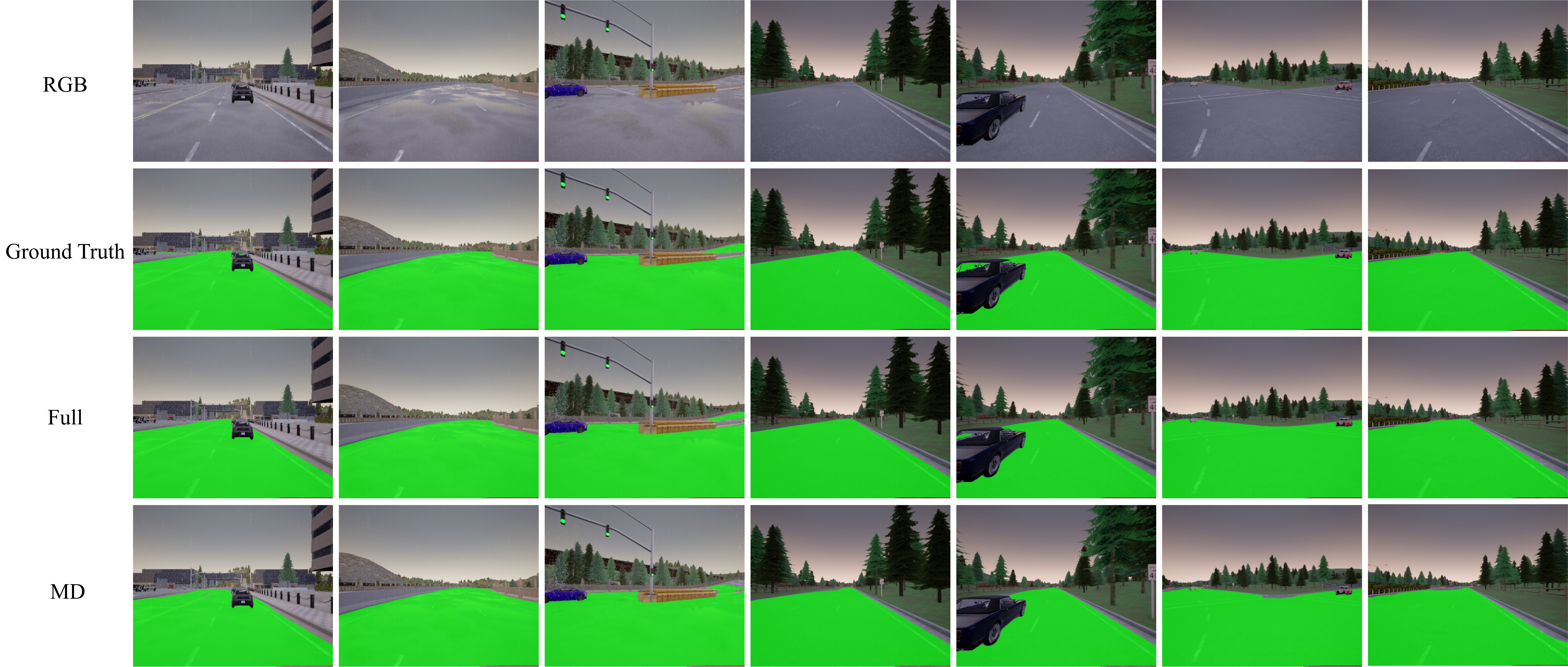}

\end{center}
  \caption{Prediction results on the R2D dataset. `Full' represents using our full framework. `MD' denotes using the proposed modality distillation method. All the images are resized for better visualization.}
\label{visualcmp_r2d}
\end{figure*}

\noindent\textbf{{Effects of Using Different Loss Weights.}}{
Because different losses are used to supervise the training of our modality distillation model, we conduct experiments to investigate the impact of using different loss weights. To reduce the complexity for our analysis, each time we only change one loss weight and fix the other two. The results of using different loss weights are shown in Fig.} \ref{loss_weight}.{ From Fig. \ref{loss_weight},
we can observe that reducing any loss weight can result in a decrease in accuracy. Therefore, three loss weights are set to equal factors of 1.0 in the modal distillation.}

\noindent\textbf{Benefits of Modality Distillation.} To reduce parameters and computational costs, we propose a modality distillation approach. The comparison between using modality distillation and without using modality distillation is shown in Table~\ref{MDSpeed}. From the comparison, we observe that the modality distillation strategy can significantly reduce the number of parameters and improve processing speed. Specifically, to generate PL, we use BTS-ResNext101 for depth estimation. 
The BTS-ResNext101 has 112.8 M parameters, and each inference takes 0.21 seconds on NVIDIA GTX 1060. However, for our road detection network, the number of parameters and the time used for each inference are about 100 M and 0.25 seconds.
Thus, the proposed  modality  distillation method can reduce nearly 50\% parameters and running time. Introducing modality distillation achieves a good trade-off between performance and computational costs.

\noindent\textbf{{Using Different Pseudo-LiDAR Representations.}} {Our framework is based on the pseudo-LiDAR generation as detailed in Section}~\ref{sec:ddif}. {Because transforming depth into 3D LiDAR points has already been used in the field of 3D object detection}~\cite{wang2019pseudo, you2019pseudo, weng2019monocular}, {we conduct experiments to investigate the performance of using pseudo-LiDAR transformation adopted by 3D object detection methods. We follow the implementation of}~\cite{wang2019pseudo} {to transform depth into the LiDAR points for a comparison. As shown in Table}~\ref{Table: pseudoLiDARCmp}, {introducing pseudo-LiDAR representation used by 3D object detection methods shows worse performance and decreases the MaxF from 96.23\% to 95.61\%}. Therefore, the pseudo-LiDAR representation for 3D object detection is not used in our road detection framework.

\subsection{Comparison with State-of-the-art Methods}

\noindent\textbf{Evaluation on the KITTI Benchmark.} We test our method on the KITTI benchmark to evaluate its performance. Some qualitative results on the testing set of KITTI benchmark are shown in Fig. \ref{KITTI_bechmark_vis}. {Precision and recall (PR) curves}~\cite{davis2006relationship, powers2020evaluation} {which show the relationship between precision and recall are obtained on the KITTI dataset}~\cite{fritsch2013new} {to evaluate the accuracy. The PR curve comparisons between our method and PLARD}~\cite{chen2019progressive} {are shown in Fig.}~\ref{pr_curves}. 
{From these curves, we can observe that for some categories, such as UU, our method can even obtain better precision if the same recall rate is used. The reason for this is that our framework is more effective for information fusion. The proposed PLIF approach in which the pseudo-LiDAR information and RGB information are used to update each other is more effective than the information fusion approach used in PLARD. We also note that our method lightly falls behind PLARD on UM. This is because the UM, UMM, and UU are combined together as a whole dataset (Urban) for training. Training the network using all the categories usually sacrifices performance in some subcategories. There is also a similar phenomenon in semantic segmentation, that is, a method obtains better overall performance, but shows slightly worse performance in some subcategories.}

We compare our models with other top ten methods published on the KITTI benchmark.
The urban road detection results which can be regarded as the overall performance of all the categories are shown in Table~\ref{Table: KITTIBenchmark}. It can be seen that our method outperforms all the other approaches on urban road detection by using only RGB images as input. Except for the overall performance, we also compare the results on three categories: UMM, UM, and UU in Table~\ref{UUUM}. From Table~\ref{UUUM}, we notice that our method obtains the best results on UU and UMM and the second best result on UM. Our method improves the best MaxF from 96.03\% to 96.63\% (+0.9\%) on UU.
To make a more intuitive comparison, a visualized metric comparison is shown in Fig.~\ref{umm_um_uu}. {From the comparison results of these three categories, we can observe that our method shows better overall performance than other methods, i.e., it outperforms other methods on most of the metrics.
Unlike previous methods which only use LiDAR information as an auxiliary to update the RGB branch, the proposed PLIF architecture enables the RGB branch and pseudo-LiDAR branch to exchange their information at different stages. Therefore, our method is more effective for information fusion to obtain better performance.}
In addition, we also compare our modality distillation results with other image-based top performing methods on the KITTI benchmark. The comparison is shown in Table~\ref{KITTIBenchmarkMod}. Benefiting from the proposed modality distillation strategy, our network can achieve highly comparable performance without using depth networks during inference.

\noindent\textbf{Evaluation on the R2D Dataset.}
We evaluate our method on the R2D dataset to demonstrate its effectiveness with different illumination and weather conditions. The visualization results of our method on the R2D dataset are shown in Fig.~\ref{visualcmp_r2d}. As can be seen from Fig.~\ref{visualcmp_r2d}, our method is robust and can obtain accurate results in different weather conditions, such as rain days and sunny days. It demonstrates that the proposed framework is effective and robust in diverse scenarios.
Similar to the comparison on the KITTI dataset,
we also use accuracy metrics to compare our method with other methods on the R2D dataset.
As the R2D dataset is a new dataset that has just been proposed in~\cite{fan2020sne}, we only compare the results which are available on the R2D dataset with ours. Unlike the KITTI dataset, the R2D dataset does not provide a sever to submit and evaluate results. Therefore, for a fair comparison, we use the evaluation code of~\cite{fan2020sne} to compute all the metrics. The detailed comparison results on the R2D dataset are shown in Table~\ref{Table: R2D}. As can be seen from Table~\ref{Table: R2D}, our method outperforms other state-of-the-art approaches on the R2D dataset by using only RGB images. 
Compared with `Ours (Full)' which takes RGB and learned PL as inputs, the modality distillation strategy slightly reduces the performance. However, the performance using modality distillation is still better than that of SNE-RoadSeg~\cite{fan2020sne}.

\section{Conclusion}

In this paper, we propose a novel pseudo-LiDAR based road detection framework. The framework is proposed to obtain state-of-the-art performance without using LiDAR. 
To achieve this goal, we use pseudo-LiDAR learned from RGB images to replace LiDAR and design a new pseudo-LiDAR based information fusion network to fuse RGB and pseudo-LiDAR information. To further optimize the proposed framework, we propose an IPPS algorithm to automatically search for the valuable information propagation paths to optimize our manually designed network. Compared with previous RGB and LiDAR-based methods, the optimized network can obtain state-of-the-art performance on the KITTI and R2D datasets without using LiDAR information. Finally, we propose a modality distillation approach which makes our method to eliminate the additional parameters and computational costs due to the generation of pseudo-LiDAR from RGB images during inference.

\section*{Acknowledgments}
We would like to express our gratitude to Tong He, Guansong Pang, and Changming Sun for their suggestions and efforts to revise the paper. We thank the editor and the reviewers for their constructive comments which help improve the quality of the paper.




\ifCLASSOPTIONcaptionsoff
  \newpage
\fi




\bibliographystyle{IEEEtran}
\bibliography{IEEEexample}

\newpage
\begin{IEEEbiographynophoto}{Libo Sun} is currently a Ph.D. candidate with the School of Computer Science, The University of Adelaide, Australia. Before that, he worked as a research associate at Nanyang Technological University, Singapore.  His current research interests include deep learning and its related applications in autonomous driving and robotics.
\end{IEEEbiographynophoto}

\begin{IEEEbiographynophoto}{Haokui Zhang} received the Ph.D. degree and the M.S. degree in computer application technology from Shannxi Provincial Key Laboratory of Speech and Image Information Processing, China, in 2021 and 2016, respectively. He is currently working as an algorithm researcher in Intellifusion, Shenzhen.
His research interests cover information retrieval, image restoration, and hyperspectral image classification.
\end{IEEEbiographynophoto}

\begin{IEEEbiographynophoto}{Wei Yin} is a Ph.D. student at School of Computer Science, The University of Adelaide,
Australia. He obtained his B.S. and M.Sc. from Nanjing University of Aeronautics and Astronautics. His research interests include monocular depth estimation and semantic segmentation. He also does some works in zero-shot transfer learning.
\end{IEEEbiographynophoto}

\end{document}